\documentclass{article}

     \usepackage[preprint]{neurips_2019}

\usepackage[utf8]{inputenc} 
\usepackage[T1]{fontenc}    
\usepackage{hyperref}       
\usepackage{url}            
\usepackage{booktabs}       
\usepackage{amsfonts}       
\usepackage{nicefrac}       
\usepackage{microtype}      
\usepackage{natbib}
\usepackage{graphicx}
\usepackage{subfig}  
\usepackage{floatrow}
\title{NeurIPS 2019 Reproducibility Challenge\\
    Ordered Memory Baselines}

\author{%
  Daniel~Borisov\\
  Department of Physiology \\
  McGill University\\
  Montreal, Canada \\
  \And
  Matthew~D’Iorio\\
  Department of Quantitative\\ Life Sciences\\
  McGill University\\
  Montreal, Canada \\
  \And
  Jeffrey~Hyacinthe\\
  Department of Quantitative\\ Life Sciences\\
  McGill University\\
  Montreal, Canada \\
}

\begin{document}

\maketitle

\begin{abstract}
  Natural language semantics can be modeled using the phrase-structured model, which can be represented using a tree-type architecture. As a result, recent advances in natural language processing have been made utilising recursive neural networks using memory models that allow them to infer tree-type representations of the input sentence sequence. These new tree models have allowed for improvements in sentiment analysis and semantic recognition. Here we review the Ordered Memory model proposed by Shen et al. (2019) at the NeurIPS 2019 conference, and try to either create baselines that can perform better or create simpler models that can perform equally as well. We found that the Ordered Memory model performs on par with the state-of-the-art models used in tree-type modelling, and performs better than simplified baselines that require fewer parameters.
\end{abstract}

\section{Introduction}
\begin{figure}[h]
  \centering
  \includegraphics[width = 5.5in]{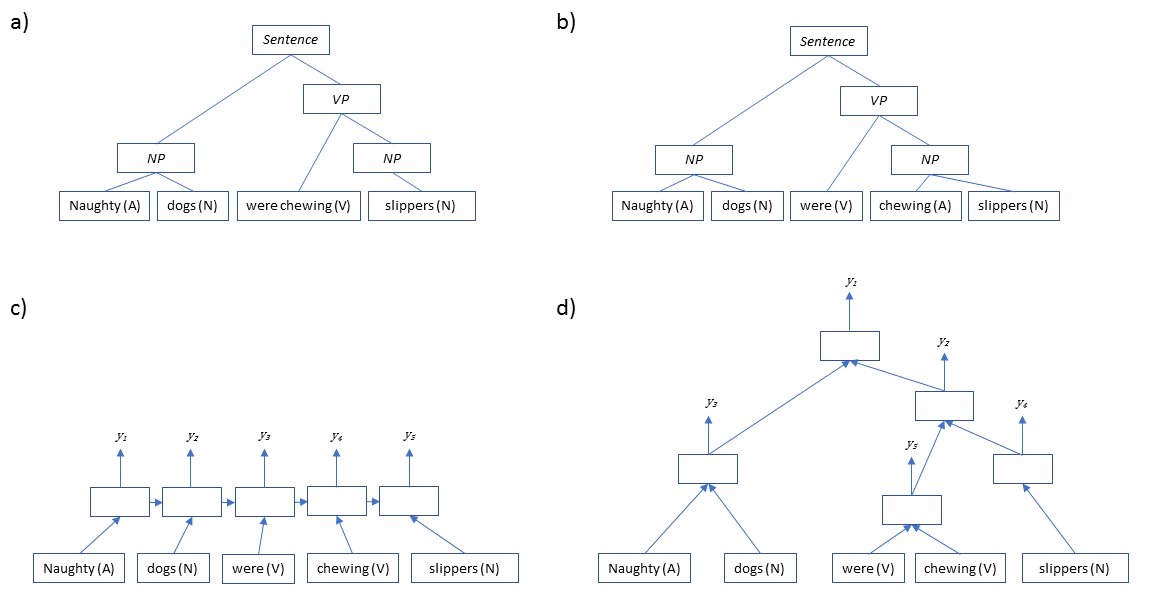}
  \caption{Phrase structure models of language using tree representations. a) Phrase structure model of a sample sentence using tree-based representation. b) Phrase structure model of a constructional homonymity, where the tree structure provides allows for the semantic differentiation of the two sentences. c) Representation of a sequential RNN model with the example sentence as an input. d) Representation of a potential inference of a tree-based RNN model with the example sentence as an input.}
  \label{fig:1}
\end{figure}

The phrase structure model is a natural language model that provides a syntactic description of a sentence by breaking it down into its constituent phrases (Chomsky, 1956). These constituent phrases, such as the Noun Phrases (NP) and Verb Phrases (VP), as well as the Verbs (V), Nouns (N), and Adjectives (A) that make them up, provide a structural analysis of a sentence and creates a mechanism for which further insight into the semantics of a specific statement can be obtained. These constituent phrases can be visually modeled utilising a tree structure, shown in Figure \ref{fig:1}a. This type of constituent analysis allows for the semantic differentiation of statements in cases of constructional homonymity, an example of which is shown in Figure \ref{fig:1}b. In this example, a native speaker would infer that the dogs were performing the act of chewing and that the sentence was not stating that the dogs were a type of slipper made for chewing. While this kind of distinction is simple for a native speaker to infer, a constituent model of the sentence can help provide the necessary information to be able to know with confidence the exact meaning of a statement.

Recurrent neural networks (RNN) were deep learning models that provided an opportunity for a network to learn the relevance of data sequences due to the recurring nature of the outputs, and provided a good mechanism for time-series data modelling, as well as providing a boost to natural language processing, due to the sequential nature of language. However, RNNs suffered from exploding and vanishing gradients, a problem which was solved by the introduction of the Long Short-Term Memory (LSTM) cell network (Hochreiter and Schmidhuber, 1997). The Gated Recurrent Unit (GRU) cell, a new cell type that hoped to solve the gradient problems of classic RNNs while providing a simpler alternative to the LSTM was also proposed (Cho et al. 2014), though suffered from a weaker performance as a result (Weiss et al. 2018). While these new models were promising for improving machine learning within the realm of natural language processing, the phrase structure model of language meant that these sequential neural networks would fail to account for the tree-structure inherent in sentences. As a result, the tree-structured LSTM (treeLSTM) was developed in order to produce a tree-architecture of LSTM cells (Tai et al. 2015). A sentence fed through a classic LSTM network would only accept the data sequentially while a treeLSTM would be able to infer the phrase structure of a sentence. The distinction is shown in Figure \ref{fig:1}c and \ref{fig:1}d, using the prior sentence example as an input to show its potential resulting representation. Even though Bowman et al. (2016) showed that sequential LSTMs can infer tree structures, and Williams et al. (2018) showed that the inferred tree structures of tree-type RNNs do not produce meaningful language structure trees, they both still conclude that tree-structured RNNs outperform sequential RNNs, and as a result, tree-based RNN architectures continue to be the dominant performing natural language semantic representation network models.

To try and accommodate for the pitfalls found within the LSTM models and their inability to produce meaningful tree structures, Shen et al. (2018) introduced a neural network architecture which they termed an Ordered Neuron LSTM (ON-LSTM), and found that this new architecture was able to produce tree structures comparable to the ground-truth trees. They expanded upon their ON-LSTM model to produce their Ordered Memory (OM) model and found that it performed on par with state-of-the-art architectures on several natural language processing datasets (Shen et al. 2019). Here we try and reproduce the baselines used as a test-case for the OM model and see if there exist models that can either perform better, or if there exist simpler models that can produce a similar result.

\section{Related Work}

Due to the many potential applications of models that can produce accurate semantic representations of natural language, there has been a lot of recent work within the field of tree-structured modelling. Memory-based recursive neural networks such as the treeLSTM have only recently been implemented within the field, and as a result there are no simple models for tree-type natural language processing. Therefore, most baselines for these classification tasks are novel implementations themselves. These include the SPINN model, the latent tree LSTM, the ON-LSTM and the treeGRU, which correspond to related work within the field. Thus, for a more detailed discussion of related works, refer to Section \ref{basemods}, where the baseline models themselves are discussed more in depth.

\section{Dataset}
\subsection{ListOps} 
\begin{figure}
    \includegraphics[width = 3in]{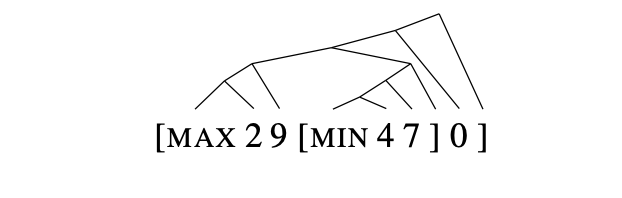}
    \vspace{-0.5cm}
    \caption{The ListOps sequenced parsed into a tree structure. Adapted from Nangia \& Bowman (2018) }
    \label{listopseg} 
\end{figure}

ListOps is a diagnostic simulated dataset introduced by Nangia \& Bowman (2018) designed to have a single parsing strategy that needs to be learned for a system to succeed at classifying it. It came following the recent work on latent tree models, a new training method for tree RNN that allows better performance than TreeRNN with conventional parsers. However, while they perform better, it was shown that they do not learn the grammar, logic or syntactic principles used for language inference.
ListOps are sequences that represent a list of operations and the values they operate on. The values are integers from 0 to 9 (modulo 10 is applied) and the class or solution for each sequence is also a single-digit integer between 0 and 9. An example of a parsed ListOps structure is shown in Figure \ref{listopseg}, where the ground truth can be solved to equal 9. 

The operators in our dataset, (train/test\_d20s)(d20s), the same one used in Ordered Memory(Min, Max, Median, Sum). We also used an alternative dataset (train/test\_d5c)(d5c) adding additional operators (First, Last, Product) to see how models performed on very similar data.
Due to resource constraints, we also used subsets of the training datasets with the 20k first examples out of the 90k original set.
An important characteristic of the ListOps dataset is that it can highlight the ability of a neural network's accuracy in parsing the syntactic structure of a stack dataset. This highlights the ability of a method to discern underlying structures that are significant in contexts of different natural language processing tasks. 

\subsection{Stanford Sentiment Treebank (SST)}

The SST dataset, created by Socher et al. (2013), is a tree-structured set of movie review sentences that have each node and tree branch labelled with a sentiment score. The SST is widely used as a baseline dataset for learning tree-structured sentences that carry sentiment data. Produced by the Stanford Natural Language Processing Group, the dataset was created to help train recursive neural network models to perform sentiment analysis. The SST was used as a sentiment-learning baseline test for the OM neural network model and as a result, was used in our experimental protocol where we attempted to use a simpler model to try and create comparable baseline results to the more complex OM model.

\section{Baseline Models}
\label{basemods}
\subsection{Model 1: Ordered Memory}
To address the local structure parsing ability of their model, Shen et al. (2019) introduced an Ordered Memory model that uses a left-to-right parsing strategy that can look one step ahead in each sequence. While similar to the recurrent latent tree models proposed by Tai et al. (2015), the OM model introduces a gated recursive cell that combines the inputs into a feed-forward network. 

Shen et al. (2019) reported a consistent result of 99.9\% accuracy. In our tests of their model, we were able to reproduce the high accuracy score with a maximum validation accuracy of 99.93\% in the best model (Table \ref{tablefull}).

\subsection{Model 2: Latent Tree LSTM}
The most accurate baseline model applied towards the ListOps dataset was the latent tree LSTM proposed by Havrylov et al. (2019). The baseline version of their tree LSTM achieved an accuracy of 66.2\% on the ListOps dataset. Their model uses the parser and a similar optimization as the  Gumbel Tree LSTM used by Choi et al. (2018). The Gumbel Tree LSTM uses gradient optimization that adds bias to the gradient estimate, and is generally not accurate towards context-free grammar tasks and the ListOps structure (Nangia \& Bowman, 2018). Havrylov et al. (2019) propose an optimization technique that aims to solve this problem with the use of a regularizing entropy term over the baseline tree distribution. 

With the optimization of the parser, addition of baseline self-critical training (Rennie et al. 2017), and Proximal Policy Optimization (PPO) (Schulman et al. 2017),  Havrylov et al. (2019) achieved a maximum accuracy of 99.2\% on the ListOps dataset. This accuracy is referenced as the top baseline compared to OM by Shen et al. (2019), however, this baseline was not reproduced in their paper. Here, we reproduce this baseline model on a subset of the ListOps model and tune hyperparameters to find its optimal application in this context.

\subsection{Model 3: SPINN}
Stack-augmented Parser-Interpreter Neural Net-work (SPINN) (Bowman et al. 2016), was introduced to address some of the issues linked to Tree-structured RNN. While conventional sequence-based RNN moves from word to word, a TreeRNN has a tree-like structure allowing a group of words to be considered together, before being separated into child nodes. This structure makes batch computation impractical. 

SPINN allows for batch computation, which provides a large speed increase over other tree-structured models and can parse unparsed data at a minimal computational cost. When evaluated on the Stanford Natural Language Inference entailment task (SNLI) (Bowman  et  al. 2015), it outperformed other models with a test accuracy of  83.2\%. However, it was one of the models that did not perform well with  the more recent ListOps dataset. It only reached an accuracy of 64.8\% while the traditional TreeLSTM could reach 98\%. These results were consistent with those found by Shen et al. (2019), and Nangia \& Bowman (2018).

\subsection{Model 4: TreeGRU}

GRU cells in RNNs tend to be quicker to train than LSTM cells due to the fewer required parameters. As a result, GRUs are used as a way to reduce computational complexity and can be trained on more resource-constrained devices. This allows for the GRU cell RNN to act as a good simple-model baseline test when comparing RNNs. However, due to the tree-structured nature of the datasets, a simple GRU would not suffice. To accommodate for this, Tsakalos and Henriques (2018) constructed a treeLSTM derivative, the treeGRU, which utilised the GRU cell in a Recursive Neural Network. This makes the treeGRU a good simple baseline test for the more complex OM model and was a baseline not considered in the original OM paper. We therefore utilised the treeGRU for the SST dataset and performed a parameter search to try and find the best performing treeGRU model.\footnote{Parameter search performed on code obtained from \url{https://github.com/VasTsak/Tree\_Structured\_GRU}} The model utilised early stopping, looking for two consecutive validation set drops, and the final model evaluated on a test set.

\section{Experimental Results}
Our experiments on an Ordered Memory, Latent TreeLSTM from Havrylov et al. (2019), and SPINN models ran on the GPU-enabled Google Colaboratory.\footnote{\url{https://colab.research.google.com}} All models were created starting from Github repository models with default parameters unless stated otherwise.
\subsection{Ordered Memory}
We were able to reproduce similar results as Shen et al. (2019) on the full original ListOps dataset. Our final test accuracy on the full dataset was 99.93\%, which is comparable to their reported result of 99.92\% accuracy (Table \ref{tablefull}). 
We also applied the model to different datasets and compared the results to other custom tuned baseline models (Table \ref{tablesub}).

\begin{figure}
    \includegraphics[width = 9cm]{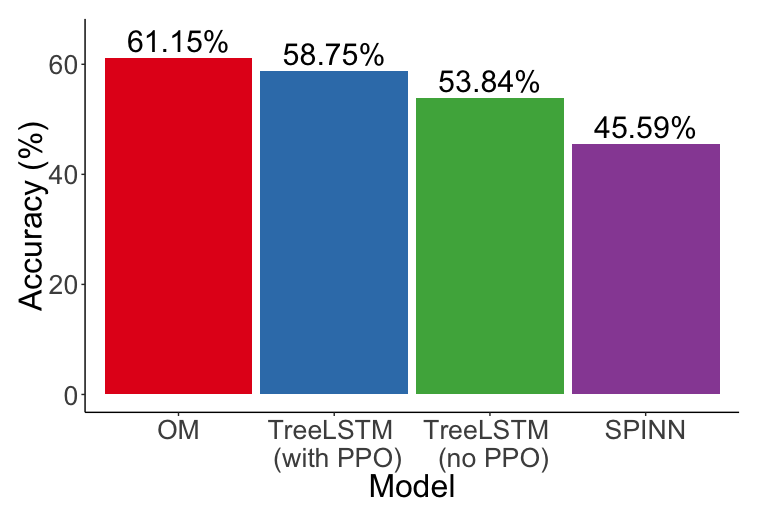}
    \vspace{0.05cm}
    \caption{The top accuracy of each model applied to the subset of the ListOps dataset. (Detailed results in Table  \ref{tablesub})}
    \label{bestss} 
\end{figure}

\subsubsection{Training set subsample}
 Ordered Memory was claimed to be data efficient due to its impressive performance on 90k examples and we were interested in exploring how it performed with even smaller training examples. With the subset of 20k examples from the originally used dataset, we found an accuracy of 61.15\% after 50 epochs. The OM model remained as the most accurate model applied to the ListOps dataset even with the much smaller dataset and baseline parameter tuning. The most accurate version of each model applied to the subset of the ListOps dataset is summarized in Figure \ref{bestss}.  

\subsubsection{Alternative dataset}
The ListOps dataset that is commonly used for baselines (test\_d20s) only had 4 operators, exploring the ListOps dataset, its generation and alternatives we found that more operators were available for potentially more complex datasets. 
With the d5c training set, subset to its first 20k examples, we obtained an accuracy of 92.21\% after 50 epochs and 97.8\% after 100 epochs (Table \ref{tablesub}). This is a very high accuracy especially as it was only a set of 20k examples and rivals the performance with train\_d20s on 90k examples and beats by a large margin the performance on 20k examples. This leads us to believe that the dataset might have been less complex, due to the new operators being easier to grasp, different in depth, or both.

\subsection{Latent Tree LSTM}
\begin{figure}[ht]
    \centering
    \subfloat[With PPO]{\includegraphics[width =6.5cm]{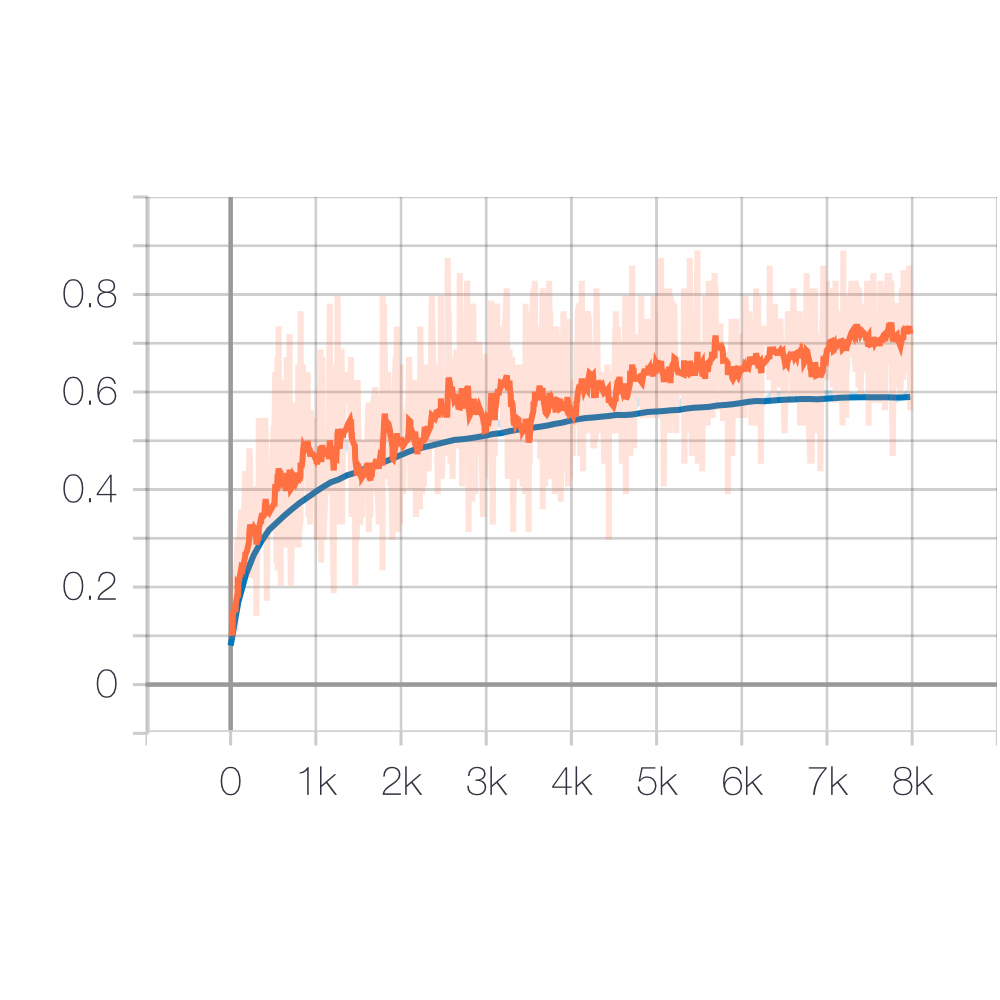}}
    \hfill
    \subfloat[Without PPO]{\includegraphics[width=6.5cm]{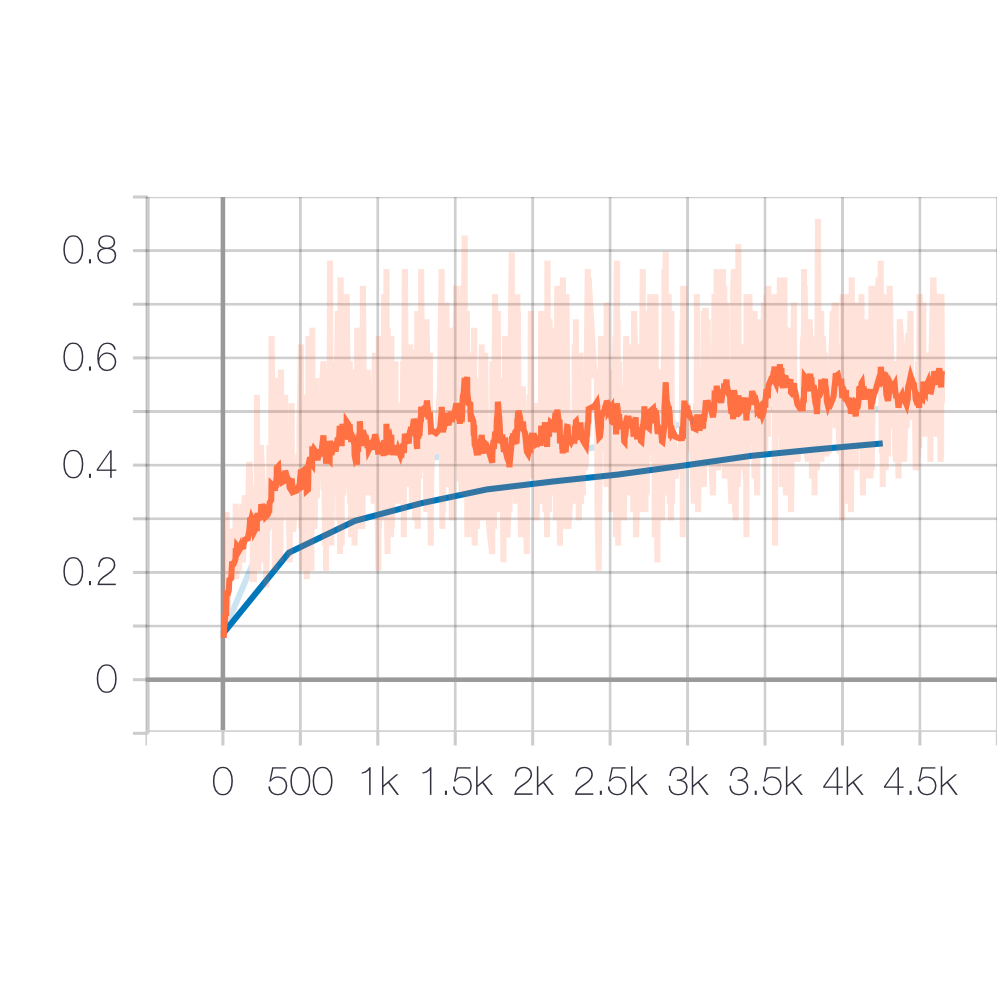}}
    \caption{Tree LSTM Model run on a subset of the ListOps, with (a) and without (b) PPO. Model accuracy (y-axis) is recorded over training iterations (x-axis). Training accuracy is recorded per batch in red and Validation accuracy is recorded at the end of each epoch in blue.}
    \label{tlstmppo} 
\end{figure}

Using a subset of the ListOps dataset, we ran the basic reinforcement model along with the PPO integrated model to observe how effectively these models learned. The training and validation accuracy over time is observed in Figure \ref{tlstmppo}. We found that the use of PPO improves the model accuracy relative to the baseline reinforcement model. The test accuracy for the tree LSTM model with PPO was 56.93\% and was 53.84\% without it.

\subsubsection{Hyperparameter tuning of latent tree lstm} 
Within the OM paper, Havrylov et al. (2019) produced the only model that could compete with their OM performances (Table \ref{tablefull}). We were thus interested in how changing the hyperparameters of the model could affect its performance. It is also worth noting that in their paper, they did not specify much in terms of how exactly they ran the algorithm, which model and with what hyperparameters. Thus we tried different hyperparameters and models from the latent TreeLSTM adapted from Havrylov et al. (2019). Since these experiments were performed on our smaller 20k training set, we also believe it can provide insight into the data efficiency of the models. 

We found that using Adam optimizer did not properly learn on our dataset, and did not properly calculate entropy. The Adadelta optimizer was the default and indeed performed much better with 53.84\% accuracy. SGD optimizer had a bit higher performance (55.82\%), but took a bit longer to run (Table \ref{tablesub}).

In terms of running time, All three optimizers had the same time spent on each epoch. Thus, the run time was more representative of the number of epochs needed to stop training than the optimizer running cost. 


When we used the alternative dataset, we found that the algorithm performed much better. In line with the higher performance of Ordered Memory, we found that Adadelta had an accuracy of 65.19\%,  Adam had 14.3\% and SGD had 62.94\% (Table \ref{tablesub}).

\begin{figure}[h]
    \centering
    \subfloat[Batch Size =32]{\includegraphics[width =4.6cm]{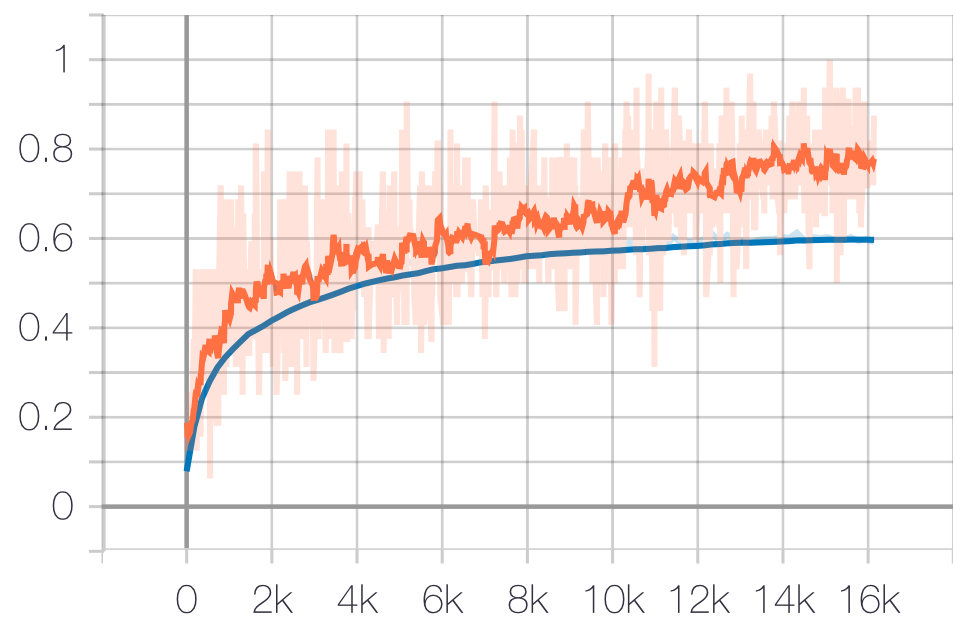}}
    \hfill
    \subfloat[Batch Size = 64]{\includegraphics[width=4.6cm]{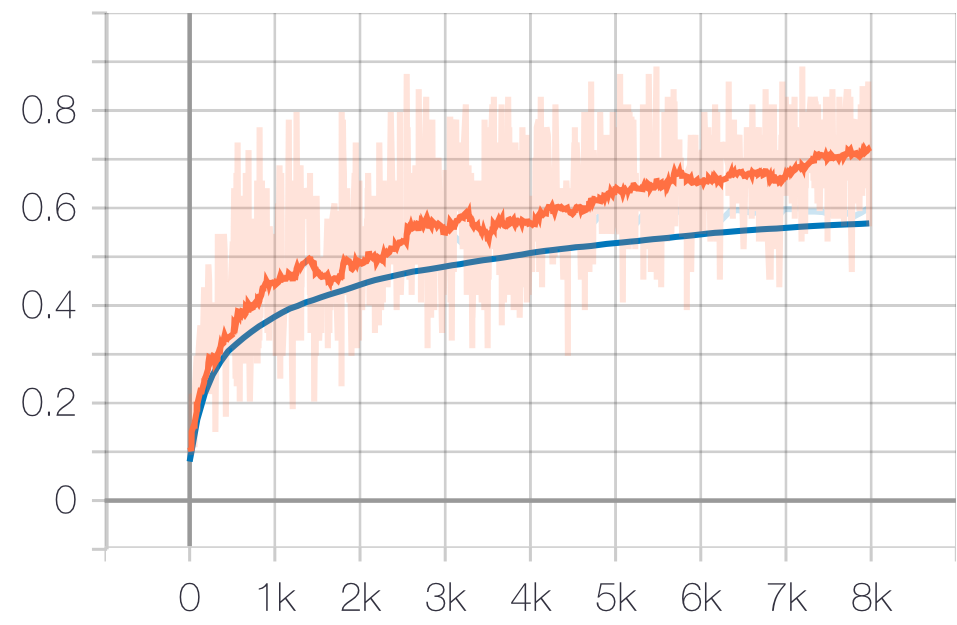}}
    \hfill
    \subfloat[Batch Size = 128]{\includegraphics[width=4.6
    cm]{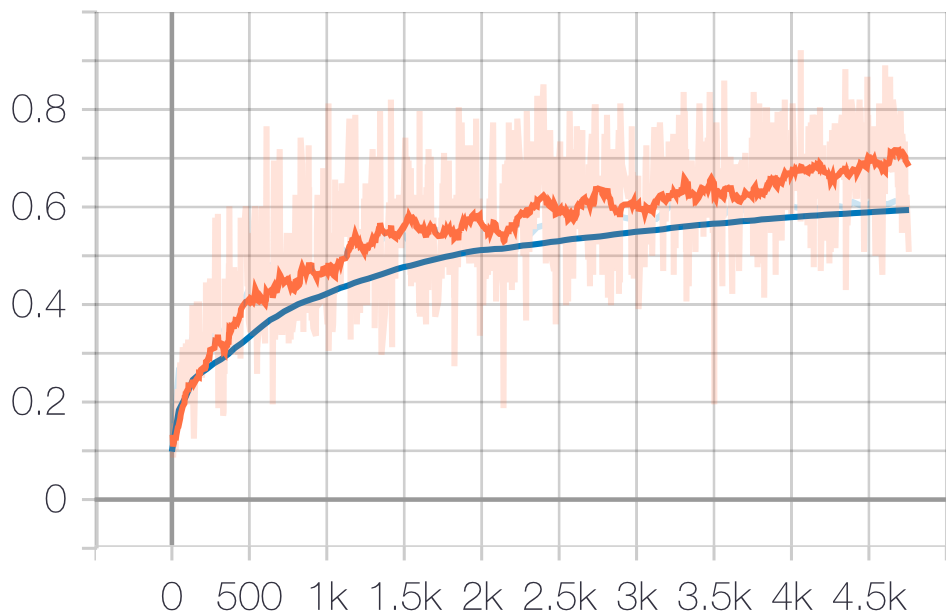}}
    \caption{Tree LSTM Model with PPO at three different batch level sizes.}
    \label{bs} 
\end{figure}

Though Havrylov et al. (2019) note that their latent Tree LSTM is robust to hyperparameter changes, we experimented with different batch sizes on the accuracy of the model. Using the superior PPO model, we compared a batch size of 32 and 128 to their default 64 value Figure \ref{bs}. Increasing the batch size from 64 to 128 increased the accuracy slightly from 53.84\% to 57.49\%, and surprisingly decreasing the batch size to 32 increased the final test accuracy the most to 58.75\%. 

\begin{table}
  \caption{Beseline and model performance reproduction (90k examples). Paper refers to Ordered Memory (Shen et al. 2018). * taken from Havrylov et al. (2019)}
  \label{tablefull}
  \centering
  \begin{tabular}{lll}
    \toprule
    Model     & Accuracy(\%)   \\
    \midrule
    Ordered memory (paper)     & 99.97  \\
    Ordered memory (ours)     & 99.93  \\
    Havrylov et al. (2019) (paper) & 99.2\\
    Havrylov et al. (2019) (without PPO)* & 64.0\\
    Havrylov et al. (2019) (without PPO) (ours) & 60.10\\
    
    \bottomrule
  \end{tabular}
\end{table}

\begin{table}[h]
  \caption{Model performance on data subset (20k examples) values below are various experiments performed on that model. Bold highlights best accuracy in experiments.}
  \label{tablesub}
  \centering
  \begin{tabular}{lllll}
    \toprule
        & Subset (d20s) & &   Alternate dataset (d5c)   \\
    \cmidrule(r){2-5}
    Model     & Accuracy(\%)     & time (min) & Accuracy(\%)     & time (min) \\
    \midrule
    Ordered Memory & \textbf{61.15}  & 65   & \textbf{97.8}  & 70     \\
    \midrule
    SPINN     & 45.6 & 555   & 22  & 56    \\
    \toprule
    \multicolumn{1}{l}{Havrylov et al. (2019)} \\
    \multicolumn{3}{l}{Optimizer}                   \\
    \midrule
    Adam & 14.98  & 24   & 14.33  & 10    \\
    adadelta     & 53.84 & 40    & \textbf{65.19} & 23    \\
    SGD     & \textbf{55.82}       & 55  & 62.94       & 10 \\
    \toprule
    \multicolumn{3}{l}{Batch size (with ppo)}                   \\
    \cmidrule(r){1-3}
    32 & \textbf{58.75}  & 373     \\
    64     & 56.93 & 330    \\
    128     & 57.49       & 125  \\
    \bottomrule
    
  \end{tabular}
\end{table}

\subsection{SPINN}
We ran the RLSPINN model on our reduced dataset and found it to have poor performance. It took much longer than the other models to run and did not learn much. On the reduced dataset the highest validation accuracy was 45.6\% in ~9h most of which was spent not improving. 

We also ran the model on the alternative dataset (d5c) and unlike the Ordered Memory and Havrylov et al. (2019), the performance was worse with 22\% accuracy (Table \ref{tablesub}).

\subsection{treeGRU}

\begin{figure}[h]
    \centering
    \subfloat[Regularization constant search, $n=5$]{\includegraphics[width =2.5in]{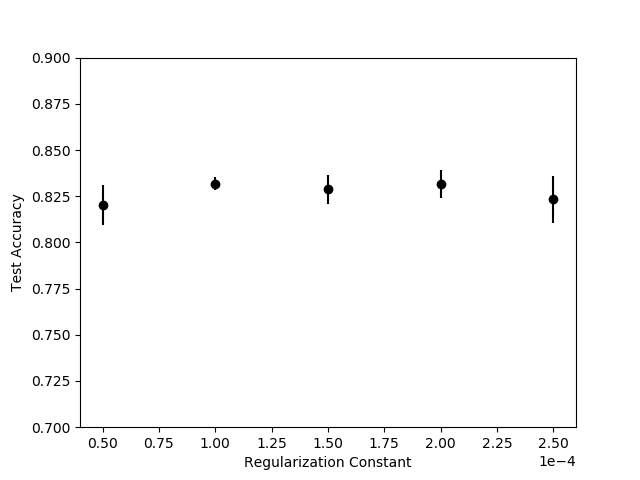}}
    \subfloat[Dropout ratio search, $n=5$]{\includegraphics[width =2.5in]{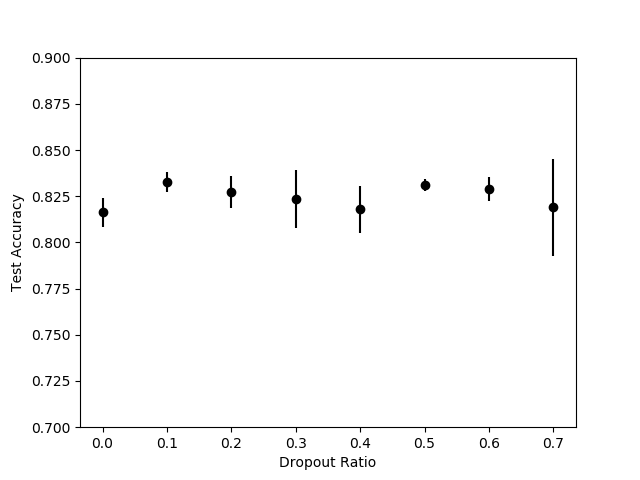}}
    \vfill
    \subfloat[Learning rate search, $n=5$]{\includegraphics[width =2.5in]{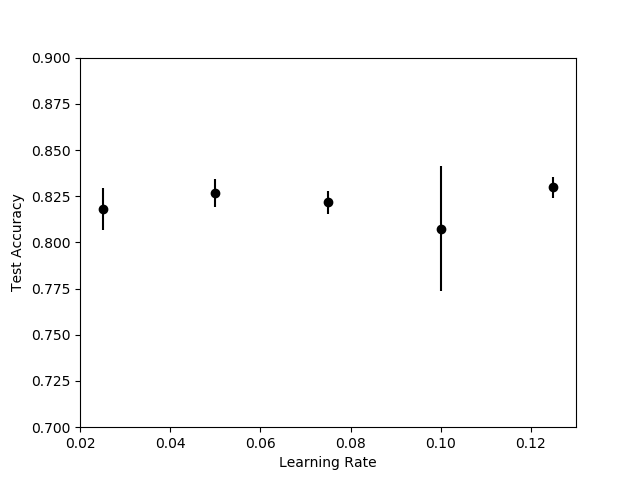}}
    \subfloat[Batch size search, $n=5$]{\includegraphics[width =2.5in]{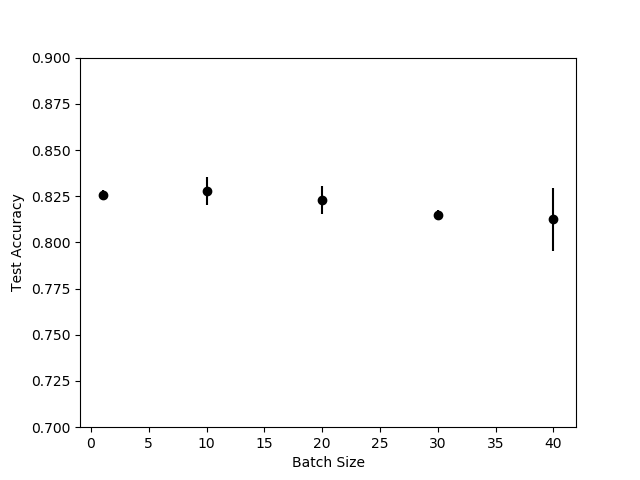}}
    \caption{Parameter search results for the treeGRU model applied to the Standford Sentiment Treebank. Vertical axes show the test accuracy.}
    \label{fig:treegrures} 
\end{figure}

A parameter search was performed on the treeGRU model, and its test-set performance after training was logged. Five training sessions were performed on each parameter value to obtain a mean and standard deviation. The parameter searched looked at the regularization constant, the dropout ratio, the learning rate, and the batch size. The results of the parameter search are summarized in Figure \ref{fig:treegrures}. While it may appear that varying the parameters changes the result, a one-way ANOVA reveals that there does not exist a statistically significant difference between the parameter variations within each of the parameter searches. The highest test accuracy obtained was 83.3\%, compared to the 90\% accuracy obtained using the OM model.

\section{Conclusion}
Of all baseline models applied to the ListOps subset, we found that the most accurate baseline other than the OM model, was a Tree LSTM adapted from Havrylov et al. (2019) The most accurate model evaluated from the Tree LSTM used a smaller batch size of 32 and their PPO enhancement. Although these changes to baseline defaults consistently slightly increased accuracy across each run, the advantages should be weighed with the cost of added time of using a smaller batch size and the PPO. 

We managed to reproduce the results of the Ordered Memory model, also reaching 99.9\% accuracy on the ListOps dataset. We wondered how the training size affected the performance and found that Even at lower training size, Ordered Memory still performs well, but only slightly better than the competitive alternatives.

We used baseline models on our subset dataset and found comparable performances for Havrylov et al. (2019) and much worse performance for SPINN. We then tried an alternative dataset containing more operators and found an increased performance in Ordered Memory and Havrylov et al. (2019) but a reduced one in SPINN. This supports that unlike SPINN, Ordered Memory and Havrylov et al. (2019) can properly interpret the syntactic structure of the input.

The treeGRU performance proved to be subpar to the OM performance, showing that the increased complexity of the OM model does produce better results than a simpler, quicker to train model.

In conclusion, we reproduced some baselines from Ordered Memory (Shen et al. 2018), validated some results and added new baselines for advances to come. The paper relied heavily on other model's stated performance, and some of the models were not clearly defined and lacking execution details. Nonetheless, The highly promising results of the Ordered Memory model were replicated and while some baselines could be improved, none of them reached the OM performance, supporting some of the conclusions of the paper.
\section*{Statement of Contribution}
Daniel set up and executed the TreeGRU experiments, Jeffrey and Matthew set up and executed the experiments on Ordered Memory, SPINN and Havrylov et al. (2019) models. Daniel contributed the Introduction and everyone contributed equally to the write-up of the report.

\section*{References}

Bowman, S.R., Angeli, G., Potts, C. and Manning, C.D., (2015). A large annotated corpus for learning natural language inference. {\it arXiv preprint arXiv:1508.05326.}

Bowman, S. R., Gauthier, J., Rastogi, A., Gupta, G., Manning, C.D.\ \&  Potts, C.\ (2016) A Fast Unified Model for Parsing and Sentence Understanding. {\it Proceedings of the 54th Annual Meeting of the Association for Computational Linguistics}

Bowman, S. R., Manning, C. D.\ \& Potts, C.\ (2015) Tree-structured composition in neural networks without tree-structured architectures. {\it arXiv preprint arXiv:1506.04834.}

Cho, K., Van Merriënboer, B., Gulcehre, C., Bahdanau, D., Bougares, F., Schwenk, H.\ \& Bengio, Y. \ (2014)  Learning phrase representations using RNN encoder-decoder for statistical machine translation. {\it arXiv preprint arXiv:1406.1078.}

Choi, J., Yoo, M.K., and Lee, S. (2018). Learning to compose task-specific tree structures.{\it Proceedings of AAAI 2018.}

Chomsky, N.\ (1956) Three models for the description of language. {\it IRE Transactions on information theory} {\bf 2}(3):113-124.

Havrylov, S., Kruszewski, G. \ \& Joulin, A., (2019). Cooperative learning of disjoint syntax and semantics. {\it arXiv preprint arXiv:1902.09393.}

Hochreiter, S.\ \& Schmidhuber, J.\ (1997) Long short-term memory. {\it Neural computation} {\bf 9}(8):1735-1780.

Nangia, N., Bowman, S.R. (2018) ListOps: A Diagnostic Dataset for Latent Tree Learning. {\it arXiv preprint arXiv:1804.06028v1}

Rennie, S.J., Marcheret, E., Mroueh, Y., 
Ross, J.,  and Goel, J. (2017). Self-critical
sequence training for image captioning. {\it Proceedings of CVPR 2017, pages 1179–1195.}

Shen, Y., Tan, S., Hosseini, A., Lin, Z., Sordoni, A.\ \& Courville, A. C. \ (2019) Ordered Memory. {\it Advances in Neural Information Processing Systems} 5038-5049.

Shen, Y., Tan, S., Sordoni, A.\ \& Courville, A.\ (2018) Ordered neurons: Integrating tree structures into recurrent neural networks. {\it arXiv preprint arXiv:1810.09536.}

Schulman, J., Wolski, F., Dhariwal, P.,
Radford, A.,  and Klimov, O. (2017). Proximal policy
optimization algorithms. CoRR, {\it arXiv preprint arXiv:1707.06347v2}

Socher, R., Perelygin, A., Wu, J., Chuang, J., Manning, C. D., Ng, A.\ \& Potts, C.\ (2013) Recursive deep models for semantic compositionality over a sentiment treebank. {\it Proceedings of the 2013 conference on empirical methods in natural language processing} 1631-1642.

Tai, K. S., Socher, R.\ \& Manning, C. D.\ (2015) Improved Semantic Representations From Tree-Structured Long Short-Term Memory Networks. {\it arXiv preprint arXiv:1503.00075.}

Tsakalos, V.\ \& Henriques, R.\ (2018) Sentiment classification using N-ary tree-structured gated recurrent unit networks. {\it Proceedings of the 10th International Joint Conference on Knowledge Discovery, Knowledge Engineering and Knowledge Management (IC3K 2018)} {\bf 1}:149-154.

Weiss, G., Goldberg, Y.\ \& Yahav, E.\ (2018) On the practical computational power of finite precision RNNs for language recognition. {\it arXiv preprint arXiv:1805.04908.}

Williams, A., Drozdov, A.\ \& Bowman, S. R.\ (2018) Do latent tree learning models identify meaningful structure in sentences?. {\it Transactions of the Association for Computational Linguistics} {\bf 6}:253-267.

\end{document}